\documentclass[10pt,letterpaper]{article}

\pdfoutput=1

\usepackage{cogsci}
\usepackage{pslatex}
\usepackage{apacite}
\usepackage{graphicx}
\usepackage{mathpazo}
\usepackage{booktabs}
\usepackage{tabularx}
\usepackage{multirow}
\newcolumntype{C}[1]{>{\centering\arraybackslash}p{#1}}

\title{Discovery and usage of joint attention in images}
 
\author{{\large \bf Daniel Harari (hararid@weizmann.ac.il)} \\
  \small{Department of Computer Science and Applied Mathematics, 
  Weizmann Institute of Science, 
  Rehovot, Israel} \\
  \small{The Center for Brains, Minds and Machines, 
  Massachusetts Institute of Technology, 
  Cambridge, MA USA} \\
  \AND
  {\large \bf Joshua B. Tenenbaum (jbt@mit.edu)} \\
  \small{Department of Brain and Cognitive Science, 
  Massachusetts Institute of Technology, 
  Cambridge, MA USA} \\
  \AND
  {\large \bf Shimon Ullman (shimon.ullman@weizmann.ac.il)} \\
  \small{Department of Computer Science and Applied Mathematics, 
  Weizmann Institute of Science, 
  Rehovot, Israel} \\
}

\begin{document}

\maketitle

\begin{abstract}
Joint visual attention is characterized by two or more individuals looking at a common target at the same time. The ability to identify joint attention in scenes, the people involved, and their common target, is fundamental to the understanding of social interactions, including others' intentions and goals. In this work we deal with the extraction of joint attention events, and the use of such events for image descriptions. 
The work makes two novel contributions. First, our extraction algorithm is the first which identifies joint visual attention in single static images. 
It computes 3-D gaze direction, identifies the gaze-target by combining gaze-direction with a 3-D depth map computed for the image, and identifies the common gaze target. Second, we use a human study to demonstrate the sensitivity of humans to joint attention, suggesting that the detection of such a configuration in an image can be useful for understanding the image, including the goals of the agents and their joint activity, and therefore can contribute to image captioning and related tasks.

\end{abstract}

\section{Introduction}
Humans, among other social species, develop remarkable capabilities for understanding social interactions in their surroundings, including others' intentions and goals \cite{Falck-Ytter2006}. In many social interactions, two or more people are engaged in joint visual attention, looking at a common target at the same time. Discovering joint attention in scenes, the people involved, and their joint target, is therefore a fundamental ability to the understanding of social interactions \cite{Moore1997}.

In this work, we first present a human survey, which demonstrates humans' sensitivity to the occurrence of joint attention in images, by comparing the ranking of image captions centred on joint-attention events, with alternative captions generated by current captioning schemes. In particular, we demonstrate the use of a joint attention event, including the participating agents and their common target, as a semantic descriptor of the image, which is shown to be more meaningful for observers, compared with alternative automatic image captions.

\begin{figure}[t]
  \centering
  \includegraphics[width=1\columnwidth]{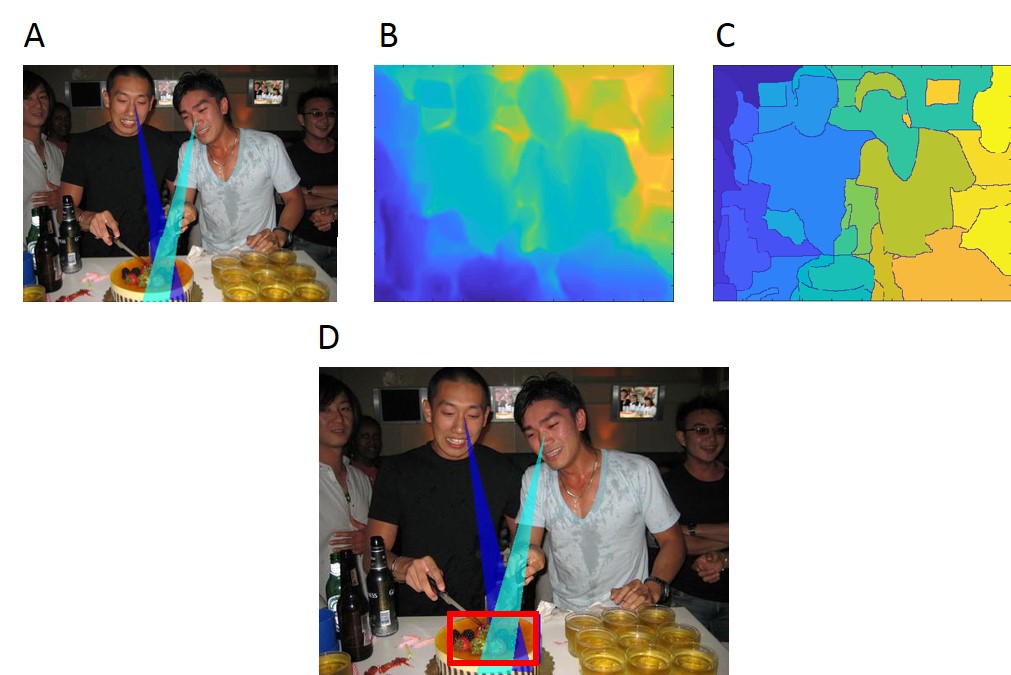}
  \caption{\small{\textbf{Discovering joint visual attention}. Our approach is compositional in the sense that different components of the process, including \textbf{(A)} face detection and 3-D gaze estimation, \textbf{(B)} depth estimation, and \textbf{(C)} image segmentation, are trained separately and then combined to perform the full task of detecting a common gaze target \textbf{(D)}.}}
\label{fig_joint_attention_components}
\end{figure}

Next, we study computationally the automatic discovery of joint visual attention in images, namely, identifying the participating agents as well as the target of joint attention. We show that deep neural networks fail to discover joint attention directly from the input images and discuss possible causes for this failure. As an alternative, we suggest a compositional approach in the sense that different components of the process are trained separately, and then combined to perform the full task (Fig.\ref{fig_joint_attention_components}). In our approach, we first detect people's faces and estimate their direction of gaze. Simple 2-D projection of the direction of gaze on the image plane is insufficient to identify gaze targets, since different objects at varying depths may lie along the 2-D line of sight, when projected onto the image plane. We therefore, use a 3-D gaze estimation model and combine the 3-D gaze direction with scene depth estimation, to identify object candidates for gaze-targets, judged by their location and depth relative to the 3-D direction of gaze. Finally, joint attention is discovered when two or more detected faces have a common gaze target. The method allows us to provide a full interpretation of the joint attention configuration, namely, identify all the participating agents as well as the target of joint attention. 

The remaining of this paper is organized as follows. Following a brief description of related work, we describe the human survey illustrating the sensitivity of human observers to joint attention events. We then describe our study on computational schemes for the discovery and interpretation of joint attention events in images, including models and experimental evaluation. Finally, we summarize and discuss our conclusions from this study.

\section{Previous work}
In the past decade, computational models of joint visual attention were introduced for a wide range of applications, from the facilitation of human-robot interactions \cite{Yucel2013}, through the detection of social interactions in videos \cite{Fathi2012}, to the discovery of joint attention signatures in child-caregiver interactions \cite{Pusiol2014}. Most of the models require the use of the scene's temporal dynamics, to infer gaze behavior and discover targets of attention \cite{Kera2016, Park2015, Pusiol2014}. For example, \citeA{Fathi2012} use videos of social interactions taken from a head-mount camera. The 3-D time-varying location of the faces around the camera wearer provides evidence for the type of social interaction in the scene, as in the case where a group of individuals look at a common location. The study by \citeA{Yucel2013} presents an image-based method for establishing joint attention between an experimenter and a robot using a video camera. A 3-D head model is used to estimate the head pose, which is then used to correct for gaze direction and object depth along the head pose direction. The estimate is further refined by a saliency-based selection to find attended objects in the video sequence. To the best of our knowledge, our work is the first to address the problem of joint attention in static 2-D images, where the temporal dynamics and 3-D information of the scene are not provided.

\section{Human sensitivity to joint attention}
To demonstrate the role of joint attention events in humans' description of images, we conducted a human survey using the Amazon Mechanical Turk platform. The survey presented natural images, each with multiple captions, and asked subjects to select their preferred caption for each image.

\subsection{Joint visual attention dataset}
\begin{figure}[t]
  \centering
  \includegraphics[width=1\columnwidth]{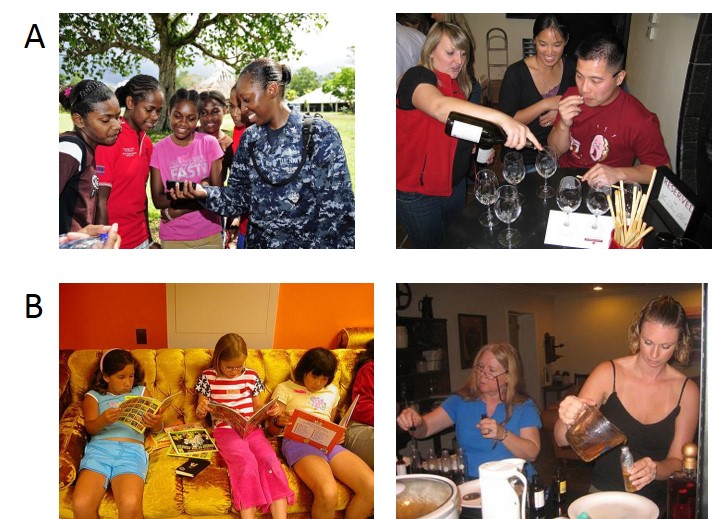}
  \caption{\small{\textbf{Joint attention dataset}. The dataset consists of 100 positive examples, where people look at a common target (row \textbf{A}), and 100 negative examples, where people attend to different targets (row \textbf{B}).}}
\label{fig_joint_attention_dataset}
\end{figure}

For the purpose of the survey and for the evaluation of compuational models as discussed in the next sections, we created a set of 200 images, extracted from the GazeFollow dataset \cite{Recasens2015}. The set includes 98 positive examples of scenes, in which people look at a common target (Fig.\ref{fig_joint_attention_dataset}A), and 102 negative examples, in which people attend to different targets (Fig.\ref{fig_joint_attention_dataset}B).

\subsection{Human study}
A subset of 60 randomly selected images were used to conduct our human survey. 53 images were positive examples of people engaged in joint visual attention. The additional 7 images were negative examples, where there was no common target, and used as a control. For each image we generated brief image captions describing the scene. One caption was an automatic 'free-style' caption generated directly from the image using a current image captioning deep neural network model \cite{Karpathy2014}. A second caption used joint attention, to generate a semantic descriptor of the image. This caption followed the simple template “\emph{X people are looking at Y}”, where \emph{X} and \emph{Y} were substituted by the number of people engaged in joint attention and the name of their common gaze target, respectively. We used our suggested scheme, which has the capability to provide a full interpretation of a joint attention event, including the participating agents and the common gaze target as described below, to generate automatically the above template-based captions. Object names were manually provided to detected targets for use in the generated captions. 

\begin{figure*}[t]
  \centering
  \includegraphics[width=1\linewidth]{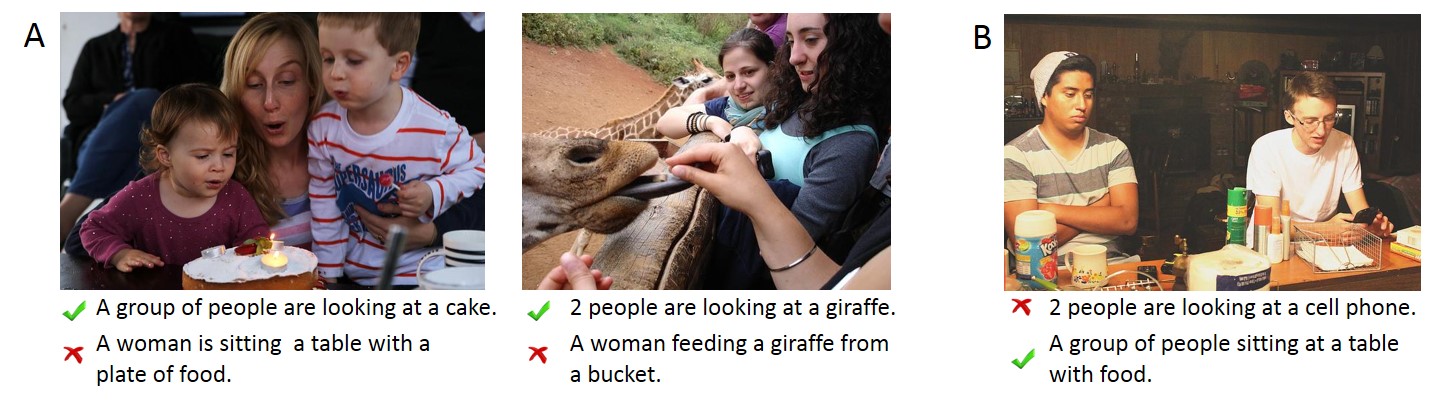}
  \caption{\small{\textbf{Use of joint attention in image captioning}. Subjects were asked to select the best matching captions for scenes. Most subjects (87\%) preferred a template caption based on detected joint attention for scenes showing people having a common gaze target \textbf{(A)}, while the alternatives (free-style or 'none') were preferred for scenes with people having different gaze targets \textbf{(B)}. }}
\label{fig_image_caption_results}
\end{figure*}

Images and captions were then presented to human subjects using the Amazon Mechanical Turk platform (25 subjects per image, 1500 total). The instructions were: \emph{"Below are several images of various scenes. For each image, use the radio buttons to mark your preference from the list of proposed captions, that best describes the image. If neither captions describe the image, select 'None'."} The list of proposed captions consisted of three alternatives at randomized order: our template-based caption, the automatic 'free-style' caption, or 'none'. \textbf{The results show that 87\% of the subjects preferred the template-based captions of our scheme}, over the automatic caption of the direct scheme, for the joint attention scenes (11\% of the subjects selected 'none' for these scenes). In contrast, only 17\% of the subjects preferred the template-based alternative for the control images, which did not present joint attention scenes (62\% of the subjects selected 'none' for these scenes). Figure \ref{fig_image_caption_results} presents some examples of the tested images along with their preferred captions. 

The results do not imply that the simplified joint-attention template is by itself an adequate caption for such images, however the strong preference over current free-style captioning (or the 'none' option) is striking. It suggests that joint attention events are highly meaningful for humans. To extract the information used in the joint-attention captions above, the joint attention model should provide a full interpretation of a joint attention event, including the participating agents and the common gaze target. The next sections describe a computational model with these capabilities.

\section{Discovery of joint attention in images}
\subsection{Direct schemes}
Deep convolutional neural networks provide a convenient framework, which is capable of learning powerful representations directly from the input data, for many visual tasks, including image classification and segmentation. For the purpose of discovering joint attention in images, we considered two alternative tasks. One was a simple binary classification, which aims to classify whether or not images include joint attention events. The second, more complex task, was to segment only common gaze target objects and not other objects. We used two netwokrs for the tasks: the VGG-16 network \cite{Simonyan2015} was adapted for the image classification task and the FCN network \cite{Long2015} for the image segmentation task, using the common practice in transfer learning \cite{goodfellow2016}. Both networks accept a single RGB image at their input. As will be shown in the evaluation section, the direct schemes fail to provide the essential interpretation of joint attention, in particular the detection of the common gaze target.

\subsection{Compositional approach}
An alternative to the direct schemes described above is a compositiona approach in the sense that different components of the process are trained separately, and then combined to perform the full task. In that sense, our method first detects gaze targets of people in an image (similar to \cite{Yucel2013}), and then determines if multiple people share a common gaze target, i.e. are engaged in joint attention. Gaze targets are detected as image segments intersecting the estimated gaze direction vector at the correct depth. 

\subsubsection{Gaze estimation} 
A key capability in our approach is the estimation of gaze directions of human agents showing in an input image. We define the direction of gaze as a 3-D vector anchored at the center location between the eyes and extending to a point in space being looked at. Several studies addressed the problem of detecting the direction of gaze in natural and unconstrained scenes, in which observed humans can look freely at targets in different directions. In a study by \citeA{Recasens2015}, a deep neural network was trained on a large dataset to infer gaze target locations in natural images directly from the input image, provided the face locations in the image. 2-D gaze directions are computed as the 2-D vectors connecting faces with their corresponding gaze targets. The reported performance of $24^{\circ}$ mean angular error of predicted gaze and a mean distance of 20\% of the image width from the correct targets, is insufficient for the discovery of common gaze targets in joint attention events. In addition, 2-D gaze direction is insufficient to discriminate between the gaze target and other objects at different depths that may lie in the line of sight, when projected onto the image plane. Other studies such as \cite{Odobez2013, Zhang2015, Zhang2017} estimate 3-D gaze direction from face and eyes images. These studies report high performance on several benchmark datasets. However, they all employ a normalization process, in which the eyes are re-rendered to obtain eye images as if the head was directly oriented at the camera. This normalization procedure sets limitations on the range of supported head poses, where large yaw and pitch angles may introduce significant distortions in the rectified eyes' images. In addition, the images in the related benchmark datasets do not include the gaze targets, which are needed to evaluate target detection from the estimated 3-D gaze directions.

In this study we use a 3-D gaze estimation model and a training dataset provided by \cite{Harari2016}. Breifly, the model, inspired by the human gaze perception mechanisms \cite{Otsuka2014, Calder2007, Langton2000}, is designed as a two-stage process, estimating the 3-D head orientation first, and the 3-D gaze direction offset from the head orientation in the second stage. In the model, image representations of the face and eyes are stored together with 3-D head orientation and 3-D gaze direction, and used later for gaze estimation using a nearest neighbours approach, which provides an efficient and convenient framework to condition the eyes' appearance of the second processing stage on the estimated head orientation from the first processing stage. In this manner, the head orientation is estimated by searching for nearest neighbours from the entire training set, while the gaze direction offset is estimated by searching for nearest neighbours only from a subset of the training set with similar head orientations. It should be noted that in contrast with other models, this model is applied to the original face and eyes, without any image transformations that may distort their appearance. 

Images of social interactions, and in particular of people engaged in joint attention, typically include the participants' upper or full body and face, as well as target objects being looked at. Such scenes in natural settings pose two major challenges for the estimation of gaze direction from the input images. First, faces, and in particular the eyes, often occupy a relatively small part of the image, and therefore appear at a limited resolution. Second, images of realistic scenes are characterized by high variability in viewing angles and scene layouts, including various face poses and target locations. \citeA{Harari2016} provide a training dataset which copes with such challenges in the input, by showing people sitting behind a table and looking freely and naturally at different target objects on and above the table at a wide range of head orientations and gaze directions.

\subsubsection{Common-gaze target detection}
The detection of a common gaze target follows the detection of individual gaze-targets. Given an input image, faces and their corresponding eyes are detected using a facial landmarks detector \cite{Baltrusaitis2013}. 3-D gaze directions are then estimated for each of the detected faces using \cite{Harari2016}. These are 3-D direction vectors whose origin are positioned at the center location between the eyes of the corresponding face (Fig.\ref{fig_joint_attention_components}A). 

In general, gaze target candidates are objects which lie in the 3-D line of sight. In an input image, we first consider for each face, gaze target candidates as segmented 2-D regions intersecting the 2-D projection of the corresponding 3-D gaze vector. We use a segmentation model \cite{Arbelaez2014} to create the 2-D object proposals (Fig.\ref{fig_joint_attention_components}C). To determine which of the 2-D object candidates intersect the 3-D gaze direction, we use a current depth estimation model \cite{Chakrabarti2016} to create a depth map of the scene (Fig.\ref{fig_joint_attention_components}B). From the depth map we extract estimated depth for the face as well as all corresponding candidate regions, as the mean depth over all image locations in those regions (depth units are in meters). Next, we locate the spatial intersection points between candidate regions and 2-D projection of the gaze direction. We estimate the depth at these points along the 3-D gaze vector by using the ratio between the X and Z components of the 3-D gaze vector and the X coordinates of the intersection points. We apply a pixel-to-meters conversion using the ratio derived from the ear-to-ear pixel distance of the face in the image and the width of an average face (0.15 meters in our implementation). The gaze target is detected as the nearest candidate region to the face, whose estimated depth matches the depth along the 3-D gaze vector at the 2-D intersection point.
Finally, joint attention is identified in an input image, when there exist a common candidate region, which is validated as the gaze target for multiple faces.

\section{Experimental evaluation}
\subsection{Testing direct schemes}
We first trained deep convolutional neural nets to discover joint attention directly from the input image. Training and testing was done using our joint attention dataset, which contains 200 images. To provide sufficient training examples, we augmented the original images, applying translations in eight directions and a horizontal flip, to yield an augmented set of 2000 images. We employ 10-fold cross validation to split the data for training and testing.

For the detection of the essential common gaze target, we trained the FCN segmentation network \cite{Long2015}, while providing annotated segmentation maps with two alternative class labels: 1=\textit{Background}, 2=\textit{Common gaze target}. Negative image examples, which do not have a common gaze target, are associated with maps with a background class label only. We evaluate the resulting segmentation in each test image by measuring the intersection over union (IOU) for the common gaze target. \textbf{The resulting mean IOU over the 200 test images was 11.8\%.} Given the limited results of the direct scheme, we also evaluated separately the stage of joint target detection, assuming that the individual directions of gaze have been extracted correctly. For this purpose, we retrained the network on images, where 2-D red lines were overlaid on each image, each line starting at a particular face and extends along the 2-D projection of the corresponding gaze direction. The rest of the training and testing was identical to the procedure described above for the original images. \textbf{The resulting mean IOU was 22.6\%}, which is double compared to the above results using the original plain images. In summary, the above network performed poorly, even when the additional information on the participating agents and gaze direction was visually available in the input, and therefore, cannot be used in an automatic system for full interpretation of joint attention events.

While direct schemes fail to capture the full interpretation of joint attention, they still may be capable of learning discriminative representations that can be used to discover and classify joint attention events. In an additional experiment, we trained the VGG-16 network \cite{Simonyan2015} to classify images as including joint attention events or not. Training and testing was similar to the above procedure used for segmentation, with the difference that a simple binary class label was provided for each of the original images instead of an annotated segmentation map. The classification accuracy over the 200 test images was 67\%. Such simplified classification task (presence vs. absence of joint attention) may be of use for some visual tasks, but it does not identify the agents and gaze target obtained by our model.

\subsection{Testing the compositional scheme}
We evaluated our compositional scheme on the 200 images of the joint attention dataset. As already mentioned, the different components of the scheme were trained separately on different datasets relevant for each component. The facial landmarks detector \cite{Baltrusaitis2013} was pre-trained on the LFPW dataset \cite{belhumeur2013}. The 3-D gaze estimation model was pre-trained on the dataset provided by \citeA{Harari2016}. The segmentation model \cite{Arbelaez2014} was pre-trained on the BSDS500 dataset \cite{Arbelaez2011}. Finally, the depth estimation model \cite{Chakrabarti2016} was pre-trained on the NYU v2 dataset \cite{silberman2012}.

We measured the performance of our scheme in interpreting joint attention scenes, by means of segmentation accuracy of the common gaze target (IOU), and detection accuracy of the participating agents. The later is defined as the percentage of the total correct detections of participating and non-participating agents out of the total number of people in the image. We tested the full scheme, which uses estimated 3D information of gaze directions and scene depth, and a reduced variant of the scheme, which does not use 3D information, but rather uses 2-D projections of the estimated gaze directions. \textbf{The 3-D scheme} yielded a mean accuracy of \textbf{67.8\%} for the detection of agents engaged in joint attention, and mean accuracy of \textbf{53.9\%} (IOU) for the segmentation of the common gaze target, which is comparable to current segmentation benchmarks \cite{Arbelaez2011}. In comparison, the 2-D variant scheme yielded a mean accuracy of 73.7\% for the detection of agents engaged in joint attention, but a much reduced mean accuracy of 33.1\% (IOU) for the segmentation of the common gaze target, demonstrating the use of 3D information in the joint attention task.

\section{Discussion}
Joint visual attention is a meaningful event, which humans are sensitive to its occurrence in an image. In the current work we studied two related aspects of joint attention: its detection in a single static image, and its contribution to image description, in particular, to automatic captioning. 
In terms of detection, the goal is to detect joint attention events, together with the participants in the events: the faces of the relevant people, their directions of gaze, and their joint target. Our algorithm is the first to produce such output based on a single image. The algorithm which worked best in our evaluations is compositional in the sense that it explicitly uses and combines several sub-components it is trained on: computing 3-D gaze direction, identifying gaze-target (by combining gaze-direction with a 3-D depth map computed for the image), and identifying the common gaze target. Our comparisons show that the different sources of information, in particular 3D gaze direction, 3D scene layout, and object segmentation, all contribute to the joint attention task. It is conceivable that with more training, a single deep network (possibly with recurrence) could be trained to extract and use all the information use for the task, but see \cite{Shalev-Shwartz2016} for possible advantages of compositional schemes in particular tasks

The sensitivity of humans to joint attention suggests that detecting such a configuration in an image can be useful for understanding the image, including the goals of the agents and their joint activity, and can therefore contribute to image captioning and related tasks. Our human study provides a partial test of this possibility.  The fixed template used for generating the joint-attention captions was highly simplified and should be replaced by more detailed and informative descriptions. However, even in this simplified form, humans often find it adequate, preferring it over the alternative captions, and rejecting it (by selecting 'none') in only a small fraction of the images. More generally, the results suggest that for automated image descriptions, it will be useful to develop methods for automatically annotating images with detected configurations which humans find particularly meaningful (including, in the use of gaze, joint attention to a target, two agents looking at each other, and others), and use them for captioning and related tasks. 

\subsubsection{Acknowledgements.}
This work was supported by the Israeli Science Foundation (ISF) grant 320/16, the German Research Foundation (DFG Grant ZO 349/1-1) and the Center for Brains, Minds and Machines (CBMM), funded by NSF STC award CCF–1231216.

\bibliographystyle{apacite}

\setlength{\bibleftmargin}{.1in}
\setlength{\bibindent}{-\bibleftmargin}

\small
\bibliography {joint_attention_arxiv}

\begin{thebibliography}{}

\bibitem [\protect \citeauthoryear {%
Arbelaez%
, Maire%
, Fowlkes%
\BCBL {}\ \BBA {} Malik%
}{%
Arbelaez%
\ \protect \BOthers {.}}{%
{\protect \APACyear {2011}}%
}]{%
Arbelaez2011}
\APACinsertmetastar {%
Arbelaez2011}%
\begin{APACrefauthors}%
Arbelaez, P.%
, Maire, M.%
, Fowlkes, C.%
\BCBL {}\ \BBA {} Malik, J.%
\end{APACrefauthors}%
\unskip\
\newblock
\APACrefYearMonthDay{2011}{}{}.
\newblock
{\BBOQ}\APACrefatitle {Contour Detection and Hierarchical Image Segmentation}
  {Contour detection and hierarchical image segmentation}.{\BBCQ}
\newblock
\APACjournalVolNumPages{IEEE TPAMI}{33}{5}{898--916}.
\PrintBackRefs{\CurrentBib}

\bibitem [\protect \citeauthoryear {%
Arbel{\'{a}}ez%
, Pont-Tuset%
, Barron%
, Marques%
\BCBL {}\ \BBA {} Malik%
}{%
Arbel{\'{a}}ez%
\ \protect \BOthers {.}}{%
{\protect \APACyear {2014}}%
}]{%
Arbelaez2014}
\APACinsertmetastar {%
Arbelaez2014}%
\begin{APACrefauthors}%
Arbel{\'{a}}ez, P.%
, Pont-Tuset, J.%
, Barron, J.%
, Marques, F.%
\BCBL {}\ \BBA {} Malik, J.%
\end{APACrefauthors}%
\unskip\
\newblock
\APACrefYearMonthDay{2014}{}{}.
\newblock
{\BBOQ}\APACrefatitle {{Multiscale combinatorial grouping}} {{Multiscale
  combinatorial grouping}}.{\BBCQ}
\newblock
\APACjournalVolNumPages{Proc of CVPR}{500}{}{328--335}.
\PrintBackRefs{\CurrentBib}

\bibitem [\protect \citeauthoryear {%
Baltru{\v{s}}aitis%
, Robinson%
\BCBL {}\ \BBA {} Morency%
}{%
Baltru{\v{s}}aitis%
\ \protect \BOthers {.}}{%
{\protect \APACyear {2013}}%
}]{%
Baltrusaitis2013}
\APACinsertmetastar {%
Baltrusaitis2013}%
\begin{APACrefauthors}%
Baltru{\v{s}}aitis, T.%
, Robinson, P.%
\BCBL {}\ \BBA {} Morency, L\BPBI P.%
\end{APACrefauthors}%
\unskip\
\newblock
\APACrefYearMonthDay{2013}{}{}.
\newblock
{\BBOQ}\APACrefatitle {{Constrained local neural fields for robust facial
  landmark detection in the wild}} {{Constrained local neural fields for robust
  facial landmark detection in the wild}}.{\BBCQ}
\newblock
\APACjournalVolNumPages{Proc of ICCV}{}{}{354--361}.
\PrintBackRefs{\CurrentBib}

\bibitem [\protect \citeauthoryear {%
Belhumeur%
, Jacobs%
, Kriegman%
\BCBL {}\ \BBA {} Kumar%
}{%
Belhumeur%
\ \protect \BOthers {.}}{%
{\protect \APACyear {2013}}%
}]{%
belhumeur2013}
\APACinsertmetastar {%
belhumeur2013}%
\begin{APACrefauthors}%
Belhumeur, P\BPBI N.%
, Jacobs, D\BPBI W.%
, Kriegman, D\BPBI J.%
\BCBL {}\ \BBA {} Kumar, N.%
\end{APACrefauthors}%
\unskip\
\newblock
\APACrefYearMonthDay{2013}{}{}.
\newblock
{\BBOQ}\APACrefatitle {Localizing parts of faces using a consensus of
  exemplars} {Localizing parts of faces using a consensus of exemplars}.{\BBCQ}
\newblock
\APACjournalVolNumPages{IEEE TPAMI}{35}{12}{2930--2940}.
\PrintBackRefs{\CurrentBib}

\bibitem [\protect \citeauthoryear {%
Calder%
\ \protect \BOthers {.}}{%
Calder%
\ \protect \BOthers {.}}{%
{\protect \APACyear {2007}}%
}]{%
Calder2007}
\APACinsertmetastar {%
Calder2007}%
\begin{APACrefauthors}%
Calder, A\BPBI J.%
, Beaver, J\BPBI D.%
, Winston, J\BPBI S.%
, Dolan, R\BPBI J.%
, Jenkins, R.%
, Eger, E.%
\BCBL {}\ \BBA {} Henson, R\BPBI N\BPBI a.%
\end{APACrefauthors}%
\unskip\
\newblock
\APACrefYearMonthDay{2007}{}{}.
\newblock
{\BBOQ}\APACrefatitle {{Separate coding of different gaze directions in the
  superior temporal sulcus and inferior parietal lobule.}} {{Separate coding of
  different gaze directions in the superior temporal sulcus and inferior
  parietal lobule.}}{\BBCQ}
\newblock
\APACjournalVolNumPages{Current Biology}{17}{1}{20--5}.
\PrintBackRefs{\CurrentBib}

\bibitem [\protect \citeauthoryear {%
Chakrabarti%
, Shao%
\BCBL {}\ \BBA {} Shakhnarovich%
}{%
Chakrabarti%
\ \protect \BOthers {.}}{%
{\protect \APACyear {2016}}%
}]{%
Chakrabarti2016}
\APACinsertmetastar {%
Chakrabarti2016}%
\begin{APACrefauthors}%
Chakrabarti, A.%
, Shao, J.%
\BCBL {}\ \BBA {} Shakhnarovich, G.%
\end{APACrefauthors}%
\unskip\
\newblock
\APACrefYearMonthDay{2016}{}{}.
\newblock
{\BBOQ}\APACrefatitle {{Depth from a Single Image by Harmonizing Overcomplete
  Local Network Predictions}} {{Depth from a Single Image by Harmonizing
  Overcomplete Local Network Predictions}}.{\BBCQ}
\newblock
\APACjournalVolNumPages{arXiv preprint arXiv:1605.07081}{}{}{}.
\PrintBackRefs{\CurrentBib}

\bibitem [\protect \citeauthoryear {%
Falck-Ytter%
, Gredeb{\"{a}}ck%
\BCBL {}\ \BBA {} von Hofsten%
}{%
Falck-Ytter%
\ \protect \BOthers {.}}{%
{\protect \APACyear {2006}}%
}]{%
Falck-Ytter2006}
\APACinsertmetastar {%
Falck-Ytter2006}%
\begin{APACrefauthors}%
Falck-Ytter, T.%
, Gredeb{\"{a}}ck, G.%
\BCBL {}\ \BBA {} von Hofsten, C.%
\end{APACrefauthors}%
\unskip\
\newblock
\APACrefYearMonthDay{2006}{}{}.
\newblock
{\BBOQ}\APACrefatitle {{Infants predict other people's action goals}} {{Infants
  predict other people's action goals}}.{\BBCQ}
\newblock
\APACjournalVolNumPages{Nature Neurosci}{9}{7}{878--879}.
\PrintBackRefs{\CurrentBib}

\bibitem [\protect \citeauthoryear {%
Fathi%
, Hodgins%
\BCBL {}\ \BBA {} Rehg%
}{%
Fathi%
\ \protect \BOthers {.}}{%
{\protect \APACyear {2012}}%
}]{%
Fathi2012}
\APACinsertmetastar {%
Fathi2012}%
\begin{APACrefauthors}%
Fathi, A.%
, Hodgins, J\BPBI K.%
\BCBL {}\ \BBA {} Rehg, J\BPBI M.%
\end{APACrefauthors}%
\unskip\
\newblock
\APACrefYearMonthDay{2012}{}{}.
\newblock
{\BBOQ}\APACrefatitle {{Social interactions: A first-person perspective}}
  {{Social interactions: A first-person perspective}}.{\BBCQ}
\newblock
\APACjournalVolNumPages{Proc of CVPR}{}{}{1226--1233}.
\PrintBackRefs{\CurrentBib}

\bibitem [\protect \citeauthoryear {%
Goodfellow%
, Bengio%
, Courville%
\BCBL {}\ \BBA {} Bengio%
}{%
Goodfellow%
\ \protect \BOthers {.}}{%
{\protect \APACyear {2016}}%
}]{%
goodfellow2016}
\APACinsertmetastar {%
goodfellow2016}%
\begin{APACrefauthors}%
Goodfellow, I.%
, Bengio, Y.%
, Courville, A.%
\BCBL {}\ \BBA {} Bengio, Y.%
\end{APACrefauthors}%
\unskip\
\newblock
\APACrefYear{2016}.
\newblock
\APACrefbtitle {Deep learning} {Deep learning}.
\newblock
\APACaddressPublisher{}{MIT press Cambridge}.
\PrintBackRefs{\CurrentBib}

\bibitem [\protect \citeauthoryear {%
Harari%
, Gao%
, Kanwisher%
, Tenenbaum%
\BCBL {}\ \BBA {} Ullman%
}{%
Harari%
\ \protect \BOthers {.}}{%
{\protect \APACyear {2016}}%
}]{%
Harari2016}
\APACinsertmetastar {%
Harari2016}%
\begin{APACrefauthors}%
Harari, D.%
, Gao, T.%
, Kanwisher, N.%
, Tenenbaum, J.%
\BCBL {}\ \BBA {} Ullman, S.%
\end{APACrefauthors}%
\unskip\
\newblock
\APACrefYearMonthDay{2016}{}{}.
\newblock
{\BBOQ}\APACrefatitle {{Measuring and modeling the perception of natural and
  unconstrained gaze in humans and machines}} {{Measuring and modeling the
  perception of natural and unconstrained gaze in humans and machines}}.{\BBCQ}
\newblock
\APACjournalVolNumPages{arXiv preprint arXiv:1611.09819}{}{}{}.
\PrintBackRefs{\CurrentBib}

\bibitem [\protect \citeauthoryear {%
Karpathy%
\ \BBA {} Fei-Fei%
}{%
Karpathy%
\ \BBA {} Fei-Fei%
}{%
{\protect \APACyear {2014}}%
}]{%
Karpathy2014}
\APACinsertmetastar {%
Karpathy2014}%
\begin{APACrefauthors}%
Karpathy, A.%
\BCBT {}\ \BBA {} Fei-Fei, L.%
\end{APACrefauthors}%
\unskip\
\newblock
\APACrefYearMonthDay{2014}{}{}.
\newblock
{\BBOQ}\APACrefatitle {{Deep Visual-Semantic Alignments for Generating Image
  Descriptions}} {{Deep Visual-Semantic Alignments for Generating Image
  Descriptions}}.{\BBCQ}
\newblock
\APACjournalVolNumPages{Proc of CVPR}{}{}{}.
\PrintBackRefs{\CurrentBib}

\bibitem [\protect \citeauthoryear {%
Kera%
}{%
Kera%
}{%
{\protect \APACyear {2016}}%
}]{%
Kera2016}
\APACinsertmetastar {%
Kera2016}%
\begin{APACrefauthors}%
Kera, H.%
\end{APACrefauthors}%
\unskip\
\newblock
\APACrefYearMonthDay{2016}{}{}.
\newblock
{\BBOQ}\APACrefatitle {{Discovering Objects of Joint Attention via First-Person
  Sensing}} {{Discovering Objects of Joint Attention via First-Person
  Sensing}}.{\BBCQ}
\newblock
\APACjournalVolNumPages{Proc of CVPR Workshops}{}{}{}.
\PrintBackRefs{\CurrentBib}

\bibitem [\protect \citeauthoryear {%
Langton%
, Watt%
\BCBL {}\ \BBA {} Bruce%
}{%
Langton%
\ \protect \BOthers {.}}{%
{\protect \APACyear {2000}}%
}]{%
Langton2000}
\APACinsertmetastar {%
Langton2000}%
\begin{APACrefauthors}%
Langton, S\BPBI R.%
, Watt, R\BPBI J.%
\BCBL {}\ \BBA {} Bruce, V.%
\end{APACrefauthors}%
\unskip\
\newblock
\APACrefYearMonthDay{2000}{}{}.
\newblock
{\BBOQ}\APACrefatitle {{Do the eyes have it? Cues to the direction of social
  attention}} {{Do the eyes have it? Cues to the direction of social
  attention}}.{\BBCQ}
\newblock
\APACjournalVolNumPages{Trends in Cog Sci}{4}{2}{50--59}.
\PrintBackRefs{\CurrentBib}

\bibitem [\protect \citeauthoryear {%
Long%
, Shelhamer%
\BCBL {}\ \BBA {} Darrell%
}{%
Long%
\ \protect \BOthers {.}}{%
{\protect \APACyear {2015}}%
}]{%
Long2015}
\APACinsertmetastar {%
Long2015}%
\begin{APACrefauthors}%
Long, J.%
, Shelhamer, E.%
\BCBL {}\ \BBA {} Darrell, T.%
\end{APACrefauthors}%
\unskip\
\newblock
\APACrefYearMonthDay{2015}{}{}.
\newblock
{\BBOQ}\APACrefatitle {{Fully Convolutional Networks for Semantic
  Segmentation}} {{Fully Convolutional Networks for Semantic
  Segmentation}}.{\BBCQ}
\newblock
\APACjournalVolNumPages{Proc of CVPR}{}{}{3431--3440}.
\PrintBackRefs{\CurrentBib}

\bibitem [\protect \citeauthoryear {%
Moore%
, Angelopoulos%
\BCBL {}\ \BBA {} Bennett%
}{%
Moore%
\ \protect \BOthers {.}}{%
{\protect \APACyear {1997}}%
}]{%
Moore1997}
\APACinsertmetastar {%
Moore1997}%
\begin{APACrefauthors}%
Moore, C.%
, Angelopoulos, M.%
\BCBL {}\ \BBA {} Bennett, P.%
\end{APACrefauthors}%
\unskip\
\newblock
\APACrefYearMonthDay{1997}{}{}.
\newblock
{\BBOQ}\APACrefatitle {{The role of movement in the development of joint visual
  attention}} {{The role of movement in the development of joint visual
  attention}}.{\BBCQ}
\newblock
\APACjournalVolNumPages{Infant Behavior and Dev}{20}{1}{83--92}.
\PrintBackRefs{\CurrentBib}

\bibitem [\protect \citeauthoryear {%
Odobez%
\ \BBA {} Mora%
}{%
Odobez%
\ \BBA {} Mora%
}{%
{\protect \APACyear {2013}}%
}]{%
Odobez2013}
\APACinsertmetastar {%
Odobez2013}%
\begin{APACrefauthors}%
Odobez, J.%
\BCBT {}\ \BBA {} Mora, K\BPBI F.%
\end{APACrefauthors}%
\unskip\
\newblock
\APACrefYearMonthDay{2013}{}{}.
\newblock
{\BBOQ}\APACrefatitle {{Person Independent 3D Gaze Estimation From Remote RGB-D
  Cameras}} {{Person Independent 3D Gaze Estimation From Remote RGB-D
  Cameras}}.{\BBCQ}
\newblock
\APACjournalVolNumPages{Proc of Int Conf on Image Proc}{}{}{}.
\PrintBackRefs{\CurrentBib}

\bibitem [\protect \citeauthoryear {%
Otsuka%
, Mareschal%
, Calder%
\BCBL {}\ \BBA {} Clifford%
}{%
Otsuka%
\ \protect \BOthers {.}}{%
{\protect \APACyear {2014}}%
}]{%
Otsuka2014}
\APACinsertmetastar {%
Otsuka2014}%
\begin{APACrefauthors}%
Otsuka, Y.%
, Mareschal, I.%
, Calder, A\BPBI J.%
\BCBL {}\ \BBA {} Clifford, C\BPBI W\BPBI G.%
\end{APACrefauthors}%
\unskip\
\newblock
\APACrefYearMonthDay{2014}{}{}.
\newblock
{\BBOQ}\APACrefatitle {{Dual-route model of the effect of head orientation on
  perceived gaze direction.}} {{Dual-route model of the effect of head
  orientation on perceived gaze direction.}}{\BBCQ}
\newblock
\APACjournalVolNumPages{J of Exp Psych: Human Perception and
  Performance}{40}{4}{1425--1439}.
\PrintBackRefs{\CurrentBib}

\bibitem [\protect \citeauthoryear {%
Park%
\ \BBA {} Shi%
}{%
Park%
\ \BBA {} Shi%
}{%
{\protect \APACyear {2015}}%
}]{%
Park2015}
\APACinsertmetastar {%
Park2015}%
\begin{APACrefauthors}%
Park, H\BPBI S.%
\BCBT {}\ \BBA {} Shi, J.%
\end{APACrefauthors}%
\unskip\
\newblock
\APACrefYearMonthDay{2015}{}{}.
\newblock
{\BBOQ}\APACrefatitle {{Social Saliency Prediction}} {{Social Saliency
  Prediction}}.{\BBCQ}
\newblock
\APACjournalVolNumPages{Proc of CVPR}{}{}{4777--4785}.
\PrintBackRefs{\CurrentBib}

\bibitem [\protect \citeauthoryear {%
Pusiol%
, Soriano%
, Fei-Fei%
\BCBL {}\ \BBA {} Frank%
}{%
Pusiol%
\ \protect \BOthers {.}}{%
{\protect \APACyear {2014}}%
}]{%
Pusiol2014}
\APACinsertmetastar {%
Pusiol2014}%
\begin{APACrefauthors}%
Pusiol, G.%
, Soriano, L.%
, Fei-Fei, L.%
\BCBL {}\ \BBA {} Frank, M.%
\end{APACrefauthors}%
\unskip\
\newblock
\APACrefYearMonthDay{2014}{}{}.
\newblock

\newblock
\APACjournalVolNumPages{Proc of Cog Sci}{}{}{}.
\PrintBackRefs{\CurrentBib}

\bibitem [\protect \citeauthoryear {%
Recasens%
, Khosla%
, Vondrick%
\BCBL {}\ \BBA {} Torralba%
}{%
Recasens%
\ \protect \BOthers {.}}{%
{\protect \APACyear {2015}}%
}]{%
Recasens2015}
\APACinsertmetastar {%
Recasens2015}%
\begin{APACrefauthors}%
Recasens, A.%
, Khosla, A.%
, Vondrick, C.%
\BCBL {}\ \BBA {} Torralba, A.%
\end{APACrefauthors}%
\unskip\
\newblock
\APACrefYearMonthDay{2015}{}{}.
\newblock
{\BBOQ}\APACrefatitle {{Where are they looking ?}} {{Where are they looking
  ?}}{\BBCQ}
\newblock
\APACjournalVolNumPages{Proc of NIPS}{}{}{1--9}.
\PrintBackRefs{\CurrentBib}

\bibitem [\protect \citeauthoryear {%
Shalev-Shwartz%
\ \BBA {} Shashua%
}{%
Shalev-Shwartz%
\ \BBA {} Shashua%
}{%
{\protect \APACyear {2016}}%
}]{%
Shalev-Shwartz2016}
\APACinsertmetastar {%
Shalev-Shwartz2016}%
\begin{APACrefauthors}%
Shalev-Shwartz, S.%
\BCBT {}\ \BBA {} Shashua, A.%
\end{APACrefauthors}%
\unskip\
\newblock
\APACrefYearMonthDay{2016}{}{}.
\newblock
{\BBOQ}\APACrefatitle {{On the Sample Complexity of End-to-end Training vs.
  Semantic Abstraction Training}} {{On the Sample Complexity of End-to-end
  Training vs. Semantic Abstraction Training}}.{\BBCQ}
\newblock
\APACjournalVolNumPages{arXiv preprint arXiv:1604.06915}{}{}{1--4}.
\PrintBackRefs{\CurrentBib}

\bibitem [\protect \citeauthoryear {%
Silberman%
, Hoiem%
, Kohli%
\BCBL {}\ \BBA {} Fergus%
}{%
Silberman%
\ \protect \BOthers {.}}{%
{\protect \APACyear {2012}}%
}]{%
silberman2012}
\APACinsertmetastar {%
silberman2012}%
\begin{APACrefauthors}%
Silberman, N.%
, Hoiem, D.%
, Kohli, P.%
\BCBL {}\ \BBA {} Fergus, R.%
\end{APACrefauthors}%
\unskip\
\newblock
\APACrefYearMonthDay{2012}{}{}.
\newblock
{\BBOQ}\APACrefatitle {Indoor segmentation and support inference from rgbd
  images} {Indoor segmentation and support inference from rgbd images}.{\BBCQ}
\newblock
\APACjournalVolNumPages{Proc of ECCV}{}{}{746--760}.
\PrintBackRefs{\CurrentBib}

\bibitem [\protect \citeauthoryear {%
Simonyan%
\ \BBA {} Zisserman%
}{%
Simonyan%
\ \BBA {} Zisserman%
}{%
{\protect \APACyear {2015}}%
}]{%
Simonyan2015}
\APACinsertmetastar {%
Simonyan2015}%
\begin{APACrefauthors}%
Simonyan, K.%
\BCBT {}\ \BBA {} Zisserman, A.%
\end{APACrefauthors}%
\unskip\
\newblock
\APACrefYearMonthDay{2015}{}{}.
\newblock
{\BBOQ}\APACrefatitle {{Very Deep Convolutional Networks for Large-Scale Image
  Recognition}} {{Very Deep Convolutional Networks for Large-Scale Image
  Recognition}}.{\BBCQ}
\newblock
\APACjournalVolNumPages{Int Conf on Learning Rep}{}{}{}.
\PrintBackRefs{\CurrentBib}

\bibitem [\protect \citeauthoryear {%
Yucel%
\ \protect \BOthers {.}}{%
Yucel%
\ \protect \BOthers {.}}{%
{\protect \APACyear {2013}}%
}]{%
Yucel2013}
\APACinsertmetastar {%
Yucel2013}%
\begin{APACrefauthors}%
Yucel, Z.%
, Salah, A\BPBI A.%
, Mericli, C.%
, Mericli, T.%
, Valenti, R.%
\BCBL {}\ \BBA {} Gevers, T.%
\end{APACrefauthors}%
\unskip\
\newblock
\APACrefYearMonthDay{2013}{}{}.
\newblock
{\BBOQ}\APACrefatitle {{Joint attention by gaze interpolation and saliency}}
  {{Joint attention by gaze interpolation and saliency}}.{\BBCQ}
\newblock
\APACjournalVolNumPages{IEEE Trans on Cybernetics}{43}{3}{829--842}.
\PrintBackRefs{\CurrentBib}

\bibitem [\protect \citeauthoryear {%
Zhang%
, Sugano%
, Fritz%
\BCBL {}\ \BBA {} Bulling%
}{%
Zhang%
\ \protect \BOthers {.}}{%
{\protect \APACyear {2015}}%
}]{%
Zhang2015}
\APACinsertmetastar {%
Zhang2015}%
\begin{APACrefauthors}%
Zhang, X.%
, Sugano, Y.%
, Fritz, M.%
\BCBL {}\ \BBA {} Bulling, A.%
\end{APACrefauthors}%
\unskip\
\newblock
\APACrefYearMonthDay{2015}{}{}.
\newblock
{\BBOQ}\APACrefatitle {{Appearance-Based Gaze Estimation in the Wild}}
  {{Appearance-Based Gaze Estimation in the Wild}}.{\BBCQ}
\newblock
\APACjournalVolNumPages{Proc of CVPR}{}{}{4511--4520}.
\PrintBackRefs{\CurrentBib}

\bibitem [\protect \citeauthoryear {%
Zhang%
, Sugano%
, Fritz%
\BCBL {}\ \BBA {} Bulling%
}{%
Zhang%
\ \protect \BOthers {.}}{%
{\protect \APACyear {2017}}%
}]{%
Zhang2017}
\APACinsertmetastar {%
Zhang2017}%
\begin{APACrefauthors}%
Zhang, X.%
, Sugano, Y.%
, Fritz, M.%
\BCBL {}\ \BBA {} Bulling, A.%
\end{APACrefauthors}%
\unskip\
\newblock
\APACrefYearMonthDay{2017}{}{}.
\newblock
{\BBOQ}\APACrefatitle {{It ' s Written All Over Your Face : Full-Face
  Appearance-Based Gaze Estimation Perceptual User Interfaces Group , Scalable
  Learning and Perception Group}} {{It ' s Written All Over Your Face :
  Full-Face Appearance-Based Gaze Estimation Perceptual User Interfaces Group ,
  Scalable Learning and Perception Group}}.{\BBCQ}
\newblock
\APACjournalVolNumPages{Proc of CVPR Workshops}{}{}{2299-2308}.
\PrintBackRefs{\CurrentBib}

\end{thebibliography}

\end{document}